\documentclass[lettersize,journal]{IEEEtran}
\usepackage{amsmath,amssymb,amsfonts}
\usepackage{array}
\usepackage{textcomp}
\usepackage{url}
\usepackage{verbatim}
\usepackage{graphicx}
\usepackage{cite}

\usepackage{booktabs}
\usepackage{algorithm}
\usepackage{algpseudocode}
\usepackage{subcaption}
\usepackage{multirow}
\usepackage{optidef}
\usepackage{bm}
\usepackage{xcolor}

\newcommand{\bbm}{\begin{bmatrix}}
	\newcommand{\ebm}{\end{bmatrix}}
\def \ssm#1{\left[ #1 \right]_{\!\times}}

\makeatletter
\newcommand*\titleheader[1]{\gdef\@titleheader{#1}}
\AtBeginDocument{%
	\let\st@red@title\@title
	\def\@title{%
		\bgroup\normalfont\large\centering\@titleheader\par\egroup
		\vskip1.5em\st@red@title}
}
\makeatother

\titleheader{This work has been submitted to the IEEE for possible publication. Copyright may be transferred without notice, after which this version may no longer be accessible.}

\title{CREVE: An Acceleration-based Constraint Approach for Robust Radar Ego-Velocity Estimation}

\author{Hoang Viet Do, Bo Sung Ko, Yong Hun Kim, and Jin Woo Song,~\IEEEmembership{Member,~IEEE}
	\thanks{Manuscript received xxx, xxx; revised xxx, xxx.
		This work was supported by Unmanned Vehicles Core Technology Research and Development Program through the National Research Foundation of Korea (NRF) (NRF-2023M3C1C1A01098408), Unmanned Vehicle Advanced Research Center (UVARC) funded by the Ministry of Science and ICT, the Republic of Korea (No. 2020M3C1C1A01086408). (Corresponding author: Jin Woo Song.)}% <-this % stops a space
	\thanks{Hoang Viet Do is with the Department of Intelligent Mechatronics Engineering and the Department of Convergence Engineering for Intelligent Drone, Sejong University, Seoul 05006, Republic of Korea (e-mail: \mbox{hoangvietdo@sju.ac.kr}).}%
	\thanks{Bo Sung Ko and Jin Woo Song are with the Department of Artificial Intelligence and Robotics and the Department of Convergence Engineering for Intelligent Drone, AIRI, Sejong University, Seoul 05006, Republic of Korea (e-mail: kobosung4756@sju.ac.kr; jwsong@sejong.ac.kr).}%
	\thanks{ Yong Hun Kim is with the Department of Artificial Intelligence and Robotics, Sejong University, Seoul 05006, Republic of Korea (e-mail: yhunkim@sejong.ac.kr).}
}

\begin{document}

\markboth{}%
{Do \MakeLowercase{\textit{et al.}}: CREVE: An Acceleration-based Constraint Approach for Robust Radar Ego-Velocity Estimation}

%\IEEEpubid{0000--0000/00\$00.00~\copyright~xxx IEEE}

\maketitle

\begin{abstract}
Ego-velocity estimation from point cloud measurements of a millimeter-wave frequency-modulated continuous wave (mmWave FMCW) radar has become a crucial component of radar-inertial odometry (RIO) systems.
Conventional approaches often exhibit poor performance when the number of outliers in the point cloud exceeds that of inliers, which can lead to degraded navigation performance, especially in RIO systems that rely on radar ego-velocity for dead reckoning.
In this paper, we propose CREVE, an acceleration-based inequality constraints filter that leverages additional measurements from an inertial measurement unit (IMU) to achieve robust ego-velocity estimations.
To further enhance accuracy and robustness against sensor errors, we introduce a practical accelerometer bias estimation method and a parameter adaptation rule that dynamically adjusts constraints based on radar point cloud inliers.
Experimental results on two open-source IRS and ColoRadar datasets demonstrate that the proposed method significantly outperforms three state-of-the-art approaches, reducing absolute trajectory error by approximately 36\%, 78\%, and 12\%, respectively.
\end{abstract}


\section{Introduction}
\begin{figure}[!t]
	\centering
	\begin{subfigure}{0.237\textwidth}
		%		\centering
		%		\includegraphics[width=1\textwidth, keepaspectratio]{Figures/radar_camera.png}
		\includegraphics[width=1.005\textwidth, height=80pt]{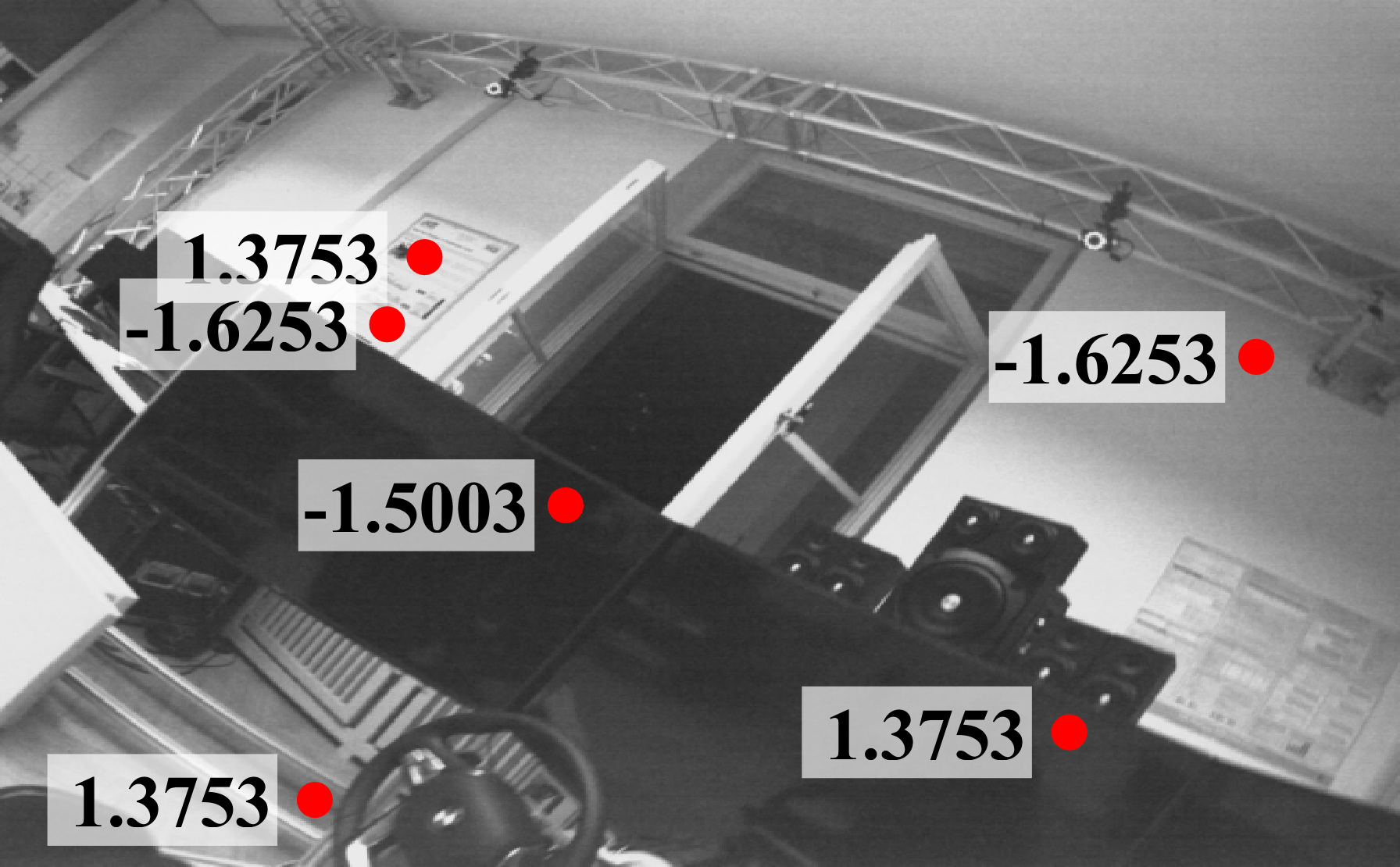}
		\caption{MoCap \texttt{Difficult}.}
		\label{introduction:figure-1-a}
	\end{subfigure}
	\begin{subfigure}{0.237\textwidth}
		%		\centering
		\includegraphics[width=1.005\textwidth, height=80pt]{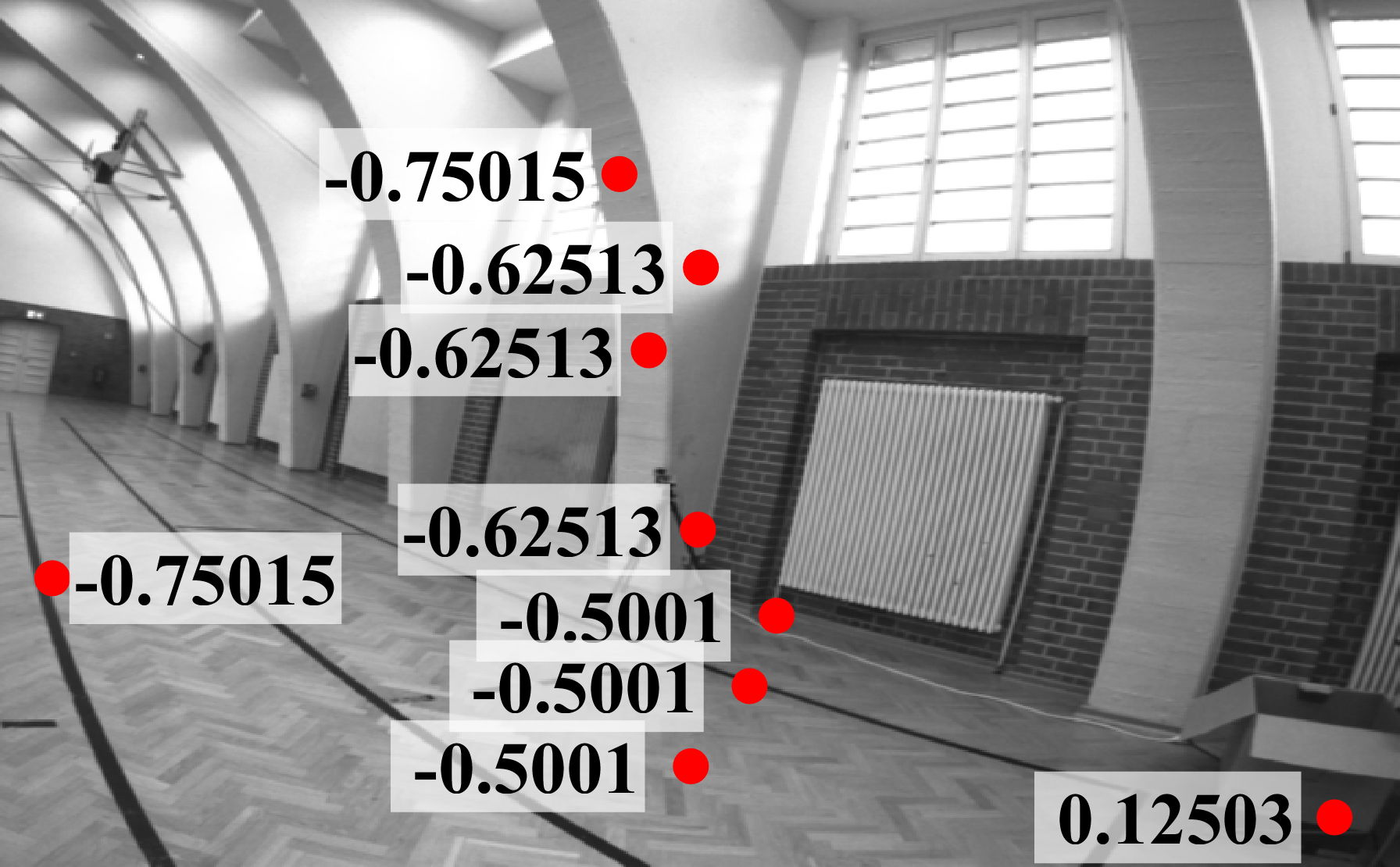}
		\caption{\texttt{Gym}.}
		\label{introduction:figure-1-b}
	\end{subfigure}
	\caption{An example of 4D point cloud measurement from an mmWave FMCW radar of the open-source IRS dataset \cite{DoerIros2021}. The radar's 3D position points are projected onto a 2D image plane (red points), with each number indicating the corresponding 1D Doppler velocity.}
	\label{introduction:figure-1}
\end{figure}
\IEEEPARstart{I}{n} recent years, millimeter-wave frequency-modulated continuous wave (mmWave FMCW) radio detection and ranging (radar) has gained significant attention in the field of robotics, particularly in applications focused on localization and navigation \cite{venon2022millimeter}.
The growing interest in radar stems from its unique advantages over traditional perception sensors such as cameras and light detection and ranging (LiDAR).
Unlike cameras, which struggle in low-light conditions or adverse weather such as rain and fog, and LiDAR, which can be hindered by environmental factors like dust or heavy precipitation, radar offers a low-cost, compact, and robust alternative capable of operating effectively in all-weather conditions \cite{visual-degraded,yang2018challenges, lee2024lidar, 8460687, 9835037}. 
This resilience makes radar particularly appealing for autonomous systems, such as drones and ground vehicles, that must navigate challenging and unpredictable environments.
Additionally, radar provides not only range and bearing information for objects of interest but also their Doppler velocity, a measurement that captures the relative motion between the sensor and its surroundings. 
In scenarios dominated by static objects, this Doppler velocity can be leveraged to estimate the ego-velocity of the sensor platform itself \cite{richards2010principles}.
Such a capability holds immense potential for enhancing the robustness and accuracy of radar-inertial odometry (RIO) systems, which combine radar data with inertial measurements to track a platform’s motion over time.

Despite these strengths, radar-based ego-velocity estimation faces significant challenges, especially when using System-on-Chip (SoC) radars.
These affordable and widely available radar systems often produce sparse and noisy point clouds due to their limited angular resolution and susceptibility to environmental interference, such as multipath reflections or clutter \cite{harlow2023new}.
This sparsity and noise make it difficult to derive reliable velocity estimates, as the data may lack the density or consistency required by traditional estimation methods.
Additionally, the substantial variation in the Doppler velocities of objects (e.g., $-1.6253$ and $1.3753$ m/s in Fig. \ref{introduction:figure-1-a}) further complicates the estimation of the sensor's velocity.
This variation can stem from the presence of dynamic objects in the environment, like moving vehicles or pedestrians, or from the sensor’s own motion across diverse terrains.
Such discrepancies make it challenging to differentiate the platform’s ego-velocity from the motion of external objects, especially in complex settings where the radar data includes a mix of static and dynamic elements.

To address the aforementioned problem, the implementation of a filtering technique or outlier removal algorithm is necessary.
This preprocessing step is crucial for subsequent odometry \cite{doer2021x} and mapping tasks \cite{4d-radar-slam}. 
Existing methods for radar-based ego-velocity estimation, such as those relying on Random Sample Consensus/Least Squares (RANSAC/LSQ) \cite{reve} have demonstrated some success but exhibit notable limitations, especially under extreme conditions \cite{drio}.
For example, when the radar point cloud contains a high proportion of outliers (e.g., measurements corrupted by noise or dynamic objects), these methods struggle to produce accurate velocity estimates.
This issue is particularly pronounced in the vertical direction ($z$-axis), where the limited elevation resolution of radar sensors provides insufficient data to resolve motion accurately \cite{dreve}.
Moreover, many conventional approaches, such as those based on feature extraction or scan matching \cite{10347455}, are designed for dense point clouds typically generated by high-end radar systems or LiDAR.
These methods falter when applied to the sparse outputs of SoC radars, restricting their applicability to specific platforms and controlled environments where dense data is available.
As a result, there is a pressing need for a robust solution that can handle the inherent sparsity and noise of SoC radar data while maintaining accuracy across diverse operational scenarios.

Generally, handling dynamic objects is considered more straightforward compared to dealing with radar ghost targets, given their arbitrary and unpredictable nature.
To tackle this, most existing radar-only research focuses on consistently extracting static objects (e.g., the ground) \cite{drio} or on using previously estimated velocities to identify unusual estimation \cite{rave}.
Nonetheless, these solutions may only be effective in specific situations and setups; for example, the radar must capture the ground for them to function properly.
This requirement is often impractical for aerial systems or narrow indoor environments.
Based on these observations, we believe that relying solely on spatial information and radar alone is insufficient.
Consequently, integrating external sensors is essential and offers a promising approach to achieving robust estimation \cite{10611471}.

In this study, we present CREVE, an acceleration-based constraint approach for robust radar ego-velocity estimation.
CREVE leverages the physical constraints derived from inertial measurement unit (IMU) acceleration measurements to bound the radar-based velocity estimates, ensuring they remain within feasible limits.
Since acceleration represents the rate of change of velocity over time, this information allows us to establish boundary values for velocity estimation.
In extreme scenarios where outliers, such as dynamic objects and ghost targets, vastly outnumber inliers, the estimation process is constrained to the acceleration surface.
This approach ensures a robust estimation solution despite the presence of such outliers.
Moreover, environmental factors such as scene changes or lack of features can be mitigated, as accelerometers are typically unaffected by these phenomena.
Although our method requires an additional IMU sensor, such sensors are commonly available in most robotics applications.
Notably, this proposed method complements our previous work in \cite{do2024dero}, which does not require accelerometers in odometry computation.
Furthermore, we enhance the robustness of the framework by incorporating a radar point cloud inlier adaptation rule to dynamically adjust constraints via a tuning parameter $\gamma$.
In our approach, the ratio of inliers to the total number of point clouds, obtained from the RANSAC/LSQ process, is continuously monitored.
When this ratio indicates that inlier measurements are nearly as numerous as the total, the constraint is intentionally relaxed by increasing $\gamma$.
This allows the system to place greater trust in the radar ego-velocity estimation when the sensor data is reliable.
Conversely, when outliers dominate the point cloud (i.e., the inlier count is low), $\gamma$ is decreased, tightening the constraint to limit the influence of potentially misleading measurements.
This adaptive mechanism enables CREVE to balance the contributions of radar and inertial data effectively, ensuring robust performance even in dynamically changing and challenging environments.

In summary, the contributions of this paper are as follows:
\begin{itemize}
	\item We propose an acceleration-based \textbf{C}onstraint method for robust \textbf{R}adar \textbf{E}go-\textbf{V}elocity \textbf{E}stimation, termed CREVE, which addresses existing radar challenges. This framework functions as a submodule within a RIO system.
	\item We present an adaptation rule to dynamically adjust the acceleration constraint bound, based on the proportion of inliers in the radar point cloud.
	\item To enhance acceleration estimation from the accelerometer, we also introduce a practical bias calculation approach that leverages two consecutive constrained estimates of ego-velocity.
	\item A comprehensive analysis is conducted against existing state-of-the-art methods using the open-source IRS \cite{DoerIros2021} and ColoRadar \cite{coloRadar} datasets. The comparison uses root mean square error and absolute trajectory error as benchmarks.
\end{itemize}

The remainder of this article is organized as follows.
Section \ref{sec:related-work} reviews related work in the context of radar-based ego-velocity estimation.
Section \ref{sec:pre} introduces the mathematical notations and provides a brief overview of traditional RANSAC and LSQ methods for ego-velocity estimation.
Building upon this foundation, Section \ref{sec:method} presents the proposed CREVE framework.
Section \ref{sec:results} evaluates the performance of the proposed method against state-of-the-art approaches using two public datasets.
Finally, Section \ref{sec:conclusion} provides concluding remarks.
\section{Related Works} \label{sec:related-work}
\subsection{Radar Ego-Velocity Estimation}
The use of radar for ego-motion estimation has been extensively explored in the literature through various methodologies.
Broadly, these approaches can be categorized into two types: 1) estimating platform ego-velocity and subsequently fusing it with data from an IMU using a filter \cite{9470842, 9649866,do2024dero}, and 2) estimating poses through scan matching algorithms \cite{8793990, 9981396,10160482}.
The former approach has attracted more attention from researchers due to the challenges posed by the sparse and limited point clouds, particularly when SoC radar is employed.
Moreover, when the focus is on odometry rather than mapping, the first approach requires only a single radar scan with a minimum of three points to estimate the platform's ego-velocity, offering significant advantages \cite{harlow2023new}.
One of the pioneering works in this field was presented by Kellner \textit{et al.} in \cite{6728341}, where a single radar, combined with a RANSAC/LSQ estimator, was used to estimate the platform's 2D ego-velocity.
Building upon this foundation, the work in \cite{6907064} employed multiple radars within the same pipeline.
Doer and Trommer \cite{reve} further advanced this field by expanding the estimation from 2D to 3D ego-velocity with the modification of the RANSAC/LSQ model from a sinusoidal in \cite{6728341} to a normalized position model of the point cloud.
In a different fashion, Park \textit{et al.} \cite{9495184} utilized two perpendicular 2D radars to derive the 3D ego-velocity.

All of the methods mentioned above share a similar foundation, employing a basic RANSAC/LSQ workflow.
However, this framework has empirically demonstrated poor performance in extreme environments, particularly in the $z$ direction, due to radar's limited elevation angle resolution \cite{harlow2023new}.
Jung \textit{et al.} \cite{dreve} improved upon this traditional approach by dividing the original workflow into a cascade of outlier removal processes for the $xy$ and $xz$ planes.
Specifically, RANSAC/LSQ is first applied to the $x$ and $y$ axes, and the resulting inliers are then fed into a second RANSAC/LSQ, focused solely on the $x$ and $z$ directions.
The authors claimed that this decoupling method improves accuracy in the $z$ direction, as the low-quality measurements from the $z$ axis do not affect the results from the $xy$ plane, thereby ensuring more reliable velocity estimation along the $x$ axis.
However, this approach increases computational cost due to the use of two RANSAC algorithms.
Alternatively, {\v{S}}tironja \textit{et al.} \cite{rave} employed a different strategy by using a sliding window of previously estimated ego-velocity values to assess the anomaly of the current estimate.
If the current outcome is identified to be unusual, it is deemed invalid and subsequently rejected.
Nevertheless, this approach reduces the number of available radar measurements, which are already limited due to the radar’s low operational frequency (typically 10 Hz).
This reduction may lead to divergence in a RIO system, where measurement updates are delayed over multiple steps
To address this issue, we project the invalid estimation onto the constraint surface, rather than discarding them completely.

Recent advancements have further expanded the scope of radar ego velocity estimation. For instance, Huang et al. \cite{10611471} introduced a physically enhanced RIO system that leverages IMU data and radar cross section (RCS) information to improve estimation robustness.
While effective, their method relies heavily on the quality of the preintegrated velocity derived from the IMU data for detecting static points.
If IMU noise and biases are not carefully accounted for, the number of valid radar points could be dramatically reduced.
In the worst-case scenario, all radar points might be rejected, thereby degrading overall system performance.
In contrast, our proposed method leverages IMU data to enhance radar ego velocity estimation quality without discarding any radar points.

\subsection{Feature-based Outlier Removal}
Unlike the previously discussed works, feature-based radar algorithms require a sufficient number of point clouds to compensate for the inherent sparsity of radar data and extract features of interest.
In this regard, using a scanning radar is more suitable than a SoC type radar.
Akarsh \textit{et al.} \cite{10161429} developed a neural network to transform radar point clouds into LiDAR-like point clouds for feature matching.
Park \textit{et al.} \cite{park2020pharao} employed phase correlation between two radar scans to infer relative motion.
In \cite{10100861}, the authors applied the graduated non-convexity method to estimate ego-velocity and perform scan-to-submap matching.
Lim \textit{et al.} \cite{orora} further refined this outcome by introducing a rotation and translation decoupling method to remove outliers for scanning radar measurement.
Alternatively, Chen \textit{et al.} \cite{drio} combined point clouds from three SoC radars to generate a sufficient density of data points, enabling the extraction of ground features from the environment.
It is important to note that all these studies primarily focus on ground vehicles in term of mapping.
On the other hand, our study aims to explore odometry across various platforms (e.g., drones) that independent of the surrounding environment.

The work by Adolfsson \textit{et al.} \cite{9636253} presents a notable feature-based approach tailored for spinning FMCW radars without Doppler information.
Their CFEAR radar odometry method employs a conservative filtering technique that retains only the k-strongest returns per azimuth, followed by the reconstruction of oriented surface points for scan matching. 
While highly efficient (running at 55 Hz on a single CPU thread) and robust across diverse environments (urban, forest, mine), CFEAR relies on dense radar scans and lacks the Doppler velocity data that our CREVE method exploits.
Similarly, Casado Herraez \textit{et al.} \cite{10610311} explore radar-only odometry and mapping, proposing two distinct approaches: a point-to-point iterative closest point (ICP) method for 3D radars and a Doppler-based constant velocity filter for 2D radars with sparse samples.
However, both approaches are validated exclusively on spinning radar systems that provide dense point clouds and rely heavily on point cloud registration algorithms.
Conversely, the proposed CREVE method is specifically designed to address the challenges of sparse radar data, such as that produced by lightweight SOC radars used in aerial robotics.
By incorporating IMU‐based acceleration constraints, CREVE achieves robust ego‐velocity estimation without relying on dense scans, thereby offering a significant advantage in terms of applicability and performance across a broad spectrum of radar sensing platforms.
\section{Preliminaries} \label{sec:pre}
\subsection{Coordinate Frame and Mathematical Notations}
In this article, the navigation (global) frame ${n}$ is defined as a local tangent frame fixed at the robot's starting point, with axes aligned to the north, east, and downward.
The IMU frame ${b}$ has axes pointing forward, right, and downward.
The FMCW radar frame ${r}$ is oriented forward, left, and upward, with its origin at the center of the transmitter antenna.
Superscripts denote the reference frame, subscripts indicate the target frame, and time indices are appended to subscripts for time-specific instances (e.g., $x_{b,k}^n$ represents a vector in ${b}$ at time step $k$, expressed in ${n}$).
The direction cosine matrix $C_b^n$ transforms vectors from frame ${b}$ to frame ${n}$ and belongs to the special orthogonal group $SO(3)$.

In mathematical notation, $\mathbb{R}^n$ represents $n$-dimensional Euclidean space, while $\mathbb{R}^{m\times n}$ refers to the set of $m \times n$ real matrices.
Scalars and vectors are written in lowercase (e.g., $x$), matrices in uppercase (e.g., $X$), with the transpose denoted as $X^{\top}$.

\subsection{RANSAC/LSQ Ego-Velocity Estimation} \label{subsec:ransac_lsq}
\begin{figure}[!t]
	\centering
	\includegraphics[scale=0.37]{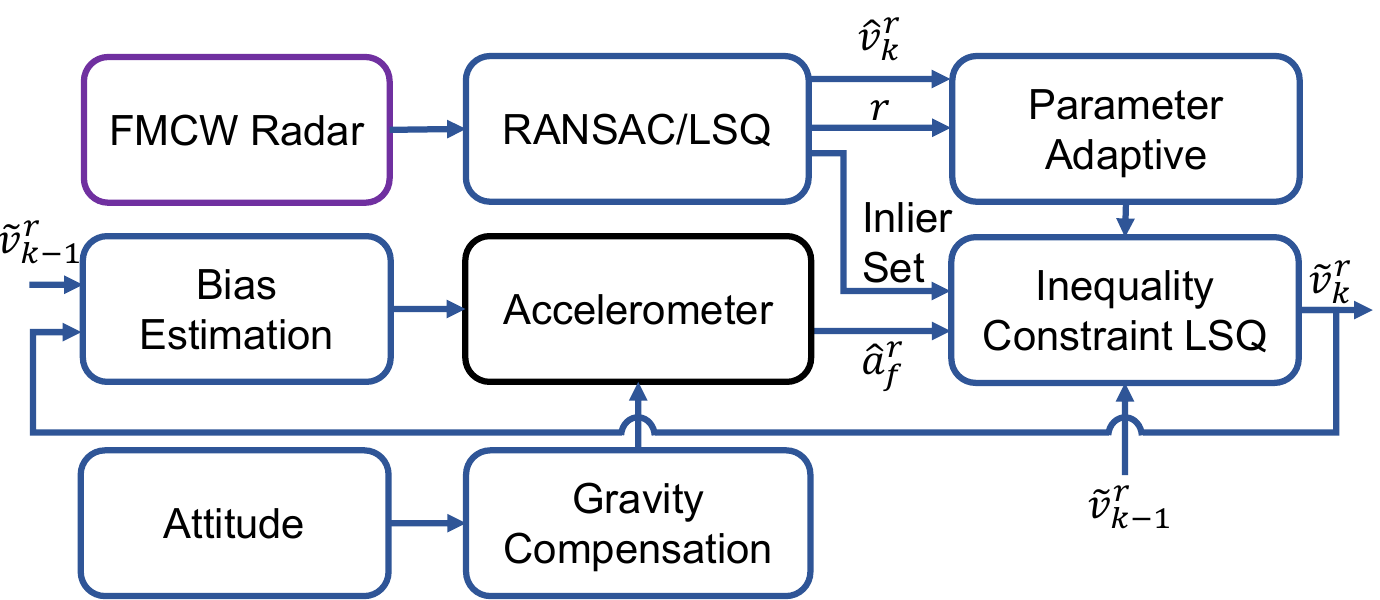}
	\caption{Block diagram overview of the proposed CREVE.}
	\label{creve_structure}
\end{figure}
In this subsection, we provide briefly the iterative RANSAC/LSQ algorithm used to estimate 3D ego-velocity from a given set of radar point clouds.
Most commercial mmWave FMCW radars produce a 4D point cloud for each spatial target, represented as $P_i = \bbm p_i  \!\!&\!\! v_{D,i} \ebm^\top$, where $p_i \in \mathbb{R}^3$ denote the position of the target $i$ expressed in $\{r\}$, and $v_{D,i}$ represents the corresponding 1D Doppler (radial) velocity.
Given $\mathcal{P}_k = \{ P_{i,k} \vert i = 1,2,\hdots,N \}$ as a set of point cloud at the time instance $k$ with $\vert \mathcal{P}_k \vert \geq 3$ (where the operator $\vert . \vert$ returns the cardinality of the set, implying $N \geq 3$ or that the set contains at least 3 points\footnote{To ensure accurate estimation, the three points must be non-coplanar.}), one can establish the following relationship \cite{reve}
\begin{align}
	\underbrace{\bbm -v_{D,1} \\ -v_{D,2} \\ \vdots \\ -v_{D,N} \ebm = \bbm \bar{p}_{x,1} & \bar{p}_{y,1} & \bar{p}_{z,1} \\ \bar{p}_{x,2} & \bar{p}_{y,2} & \bar{p}_{z,2} \\  \vdots & \vdots & \vdots \\ \bar{p}_{x,N} & \bar{p}_{y,N} & \bar{p}_{z,N} \ebm \bbm v_x^r \\ v_y^r \\ v_z^r \ebm}_{y_k = H_k v_k^r}.
\end{align}
Here, $v_k^r \in \mathbb{R}^3$ is the radar velocity vector at time step $k$, and the direction vector $\bar{p}_i$ of target $i$ is obtained by normalizing $\bar{p}_i$, such that $\bar{p}_i = p_i / \Vert p_i \Vert_2^{\vphantom{2}}$.
The estimate radar velocity $\hat{v}_k^r$ can be obtained by minimizing the squared $2$-norm loss function $\hat{v}_k^r = \mathrm{arg\,min} \frac{1}{2}\big\Vert H_k v_k^r - y_k \big\Vert_2^2$ .
The well-known solution, given in \cite{simon2006optimal}, is $\hat{v}_k^r = \big( H_k^\top H_k\big)^{-1} H_k^\top y_k$.

The result discussed above does not account for outliers.
To effectively handle outliers, the widely-used iterative RANSAC method \cite{ransac} can be employed.
Specifically, in each iterative, out of a fixed number of total iterations, three points from $\mathcal{P}_k$ are randomly select to compute a temporary estimate of $\hat{v}^r_k$.
Inlier points are then determined by caclulating the absolute error between $y_k$ and $H_k \hat{v}_r^k$ for all points in $\mathcal{P}_k$, then comparing the error to a pre-determined threshold.
Finally, the estimate $\hat{v}^r_k$ is refined by re-computing it using the inlier set.
\section{Methodology} \label{sec:method}
\subsection{Overview}
We begin by presenting an overview of our CREVE system, as depicted in the block diagram in Fig. \ref{creve_structure}.
First, the radar point cloud is processed using a traditional RANSAC/LSQ algorithm to extract an inlier set based on a prior estimate of the radar ego velocity.
During this step, the ratio of inliers to the total number of points in the radar point cloud is computed and subsequently used to determine the adaptive parameter $\gamma$.
Next, the platform's acceleration is derived from the accelerometer output.
This process incorporates our proposed bias estimation, which is obtained from two consecutive radar ego velocity estimations, and gravity compensation that is facilitated by gyroscope measurements.
The resulting acceleration and the adaptive parameter $\gamma$ are then used to formulate the inequality constraints.
Finally, the inlier set and these inequality constraints are integrated into the proposed acceleration constrained least squares optimization.
Solving this optimization problem yields the final, robust estimation of the radar ego velocity.

\subsection{Acceleration-constrained Least Squares} \label{subsection:lsq}
Given the rotation matrix and accelerometer bias estimation $\hat{b}_a$ at time instance $k$, the estimated acceleration expressed in the radar frame $\hat{a}_f^r$ can be calculated as
\begin{align} \label{a_cal}
	\hat{a}_f^r = C_b^r \big( \bar{f}^b - \hat{b}_a + C_n^b g^n \big).
\end{align}
Here, $g^n$ represents the gravitational force vector, defined as $g^n = \bbm 0 & 0 & g \ebm^\top$, where $g$ is the gravitational constant.
The rotation matrix (extrinsic parameter) between the IMU and the radar $C_b^r$ is assumed to be given.
%The symbol $C_b^r$ denotes the rotation matrix between the IMU and the radar, which can be obtained through extrinsic calibration process.
%
For the sensor modeling, we adopt the model introduced in \cite{do2024dero}, which is given by $\bar{f}^b = C_n^b (a^n - g^n) + b_a + n_a$.
In this expression, $a^n$ and $n_a$ respectively signify the acceleration in the navigation frame and accelerometer noise.

Based on \eqref{a_cal}, we formulate the following linear programming (LP) problem
\begin{argmini!} 
	{v_k^r \in \mathbb{R}^3}{\frac{1}{2} \big\Vert H_k v_k^r - y_k \big\Vert_2^2}
	{\label{lp}}{\tilde{v}_k^r=}
	\addConstraint{{v}^r_k - {v}^r_{k-1}\geq \hat{a}^r_{f, k} \Delta t_r - \gamma_k \label{lp_1}}
	\addConstraint{{v}^r_k - {v}^r_{k-1}\leq \hat{a}^r_{f, k} \Delta t_r + \gamma_k. \label{lp_2}}
\end{argmini!}
%\begin{mini!}|l|[]
%	{\tilde{v}_k^r \in \mathbb{R}^3}{\frac{1}{2} \big\Vert H_k \tilde{v}_k^r - y_k \big\Vert_2^2}
%	{\label{lp}}{}
%	\addConstraint{\tilde{v}^r_k \geq \big( \hat{a}^r_{f, k} - \gamma \big) \Delta t_r + \tilde{v}^r_{k-1}}
%	\addConstraint{\tilde{v}^r_k \leq \big( \hat{a}^r_{f, k} + \gamma \big) \Delta t_r + \tilde{v}^r_{k-1}.}
%\end{mini!}
%%
In this relation, $\tilde{v}_k^r \in \mathbb{R}^3$ indicates the constrained ego-velocity estimation at time $k$, and $\Delta t_r$ is the radar time period.
The LP above follows a standard form \cite{ineqlsq} and can be solved using various existing iterative methods.
The inequality constraint is derived from the approximate relationship between average acceleration and velocity, given by $a^r_k = (v^r_k - v^r_{k-1}) / \Delta t_r$.
It is essential to recognize that these inequalities are subject to various sources of error, including attitude and bias estimation errors from \eqref{a_cal} as well as discretization error.
To take this issue into account, we introduce a tuning parameter $\gamma \in \mathbb{R}^3$.
Incorporating $\gamma$ ensures that the nominal is adequately represented in the LP.
Also, this parameter allows for the adjustment of the constraint range, providing flexibility to either tighten or loosen the constraints as needed.
\begin{figure}[t]
	\centering
	\includegraphics[scale=0.185]{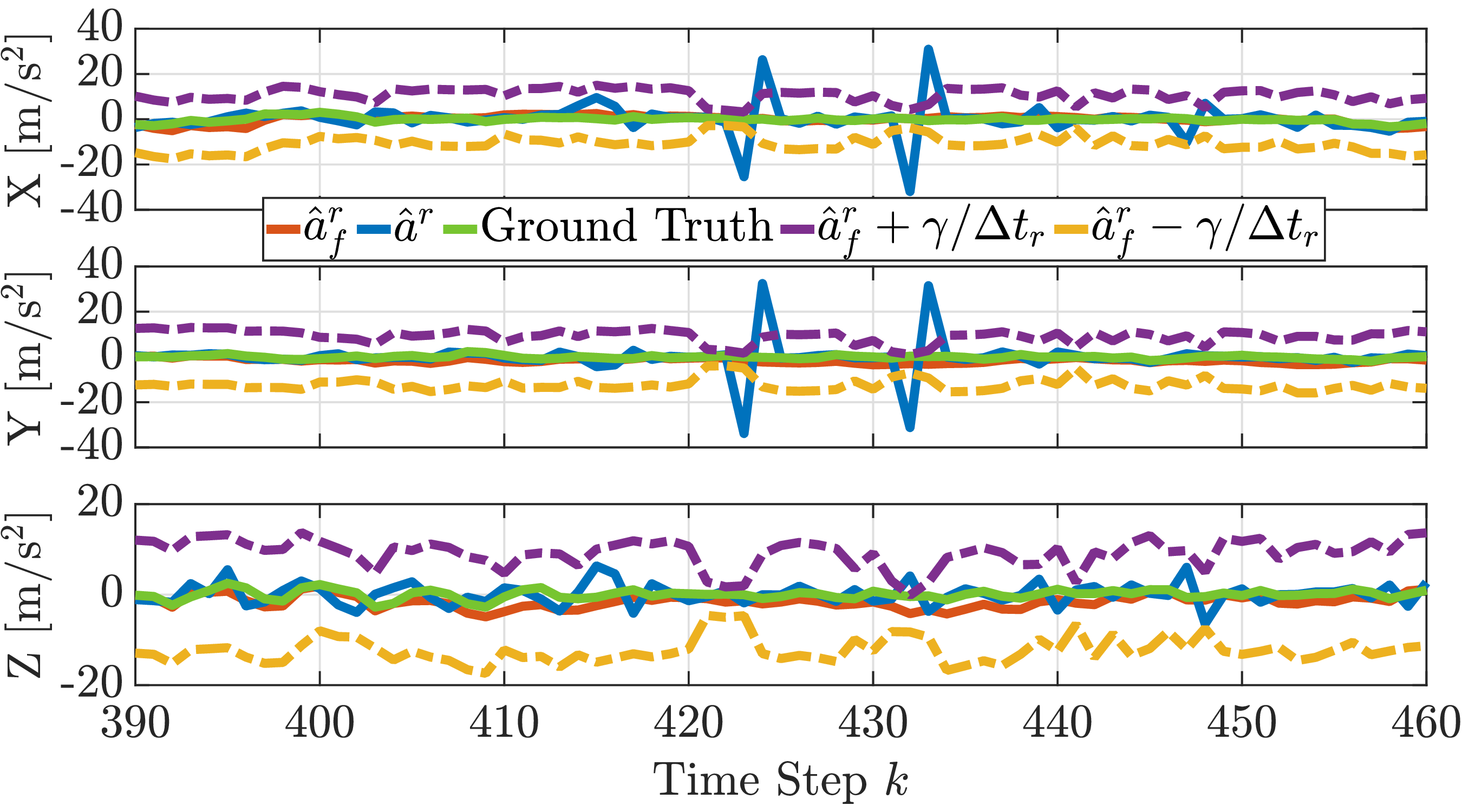}
	\caption{Our idea visualization using the IRS dataset \cite{DoerIros2021}, comparing the estimated $\hat{a}_f^r$ calculated from \eqref{a_cal}, MoCap ground truth, and average acceleration derived from $\hat{v}^r$. For this example, we used $\gamma_{\textrm{min}} = \bbm 0.05 \!\!&\!\! 0.05 \!\!&\!\! 0.05 \ebm^\top$ m/s and $\gamma_{\textrm{max}} = \bbm 1.25 \!\!&\!\! 1.25 \!\!&\!\! 1.25 \ebm^\top$ m/s (best viewed in color).}
	\label{proposed_check}
\end{figure}

Our approach is visually demonstrated in Fig. \ref{proposed_check}.
After compensating for bias and gravity, the estimated acceleration $\hat{a}^r$ provides valuable insight into the relationship between the current and previous velocity estimates.
Additionally, the effectiveness of our proposed tuning parameter $\gamma$ is evident; when appropriately selected, it ensures that the ground truth acceleration consistently remains within the prescribed constraint boundaries.
It is worth mentioning that we applied the standard method from \cite{reve} to produce this example, further supporting our claim that this method is not robust to outliers, even in a static environment, as evidenced by the peaks exceeding the constraint range.

\subsection{Inlier-based Adaptation Rule}
As shown above, the value of the adaptive parameter $\gamma$ plays a critical role in our approach.
If $\gamma$ is too large, the CREVE system may incorporate inaccurate estimations (see Fig. \ref{proposed_check}).
In contrast, If $\gamma$ is too small, the radar’s velocity measurements might be effectively ignored due to the imposed inequality constraints \eqref{lp_1} and \eqref{lp_2}.
In the worst-case scenario, the optimization problem \eqref{lp} could even become infeasible.
The question is how to choose the right value for this parameter.
Therefore, determining an appropriate value for $\gamma$ is crucial.
In practice, choosing a suitable $\gamma$ for general-purpose applications can be challenging.
For a specific platform (e.g., a drone or a ground vehicle),  $\gamma$ could be chosen empirically because the platform’s unique characteristics provide useful guidance.
However, since our study is not targeted at any specific platform, we address the tuning challenge by proposing an adaptive rule that leverages the radar inlier point cloud generated by the traditional RANSAC/LSQ method.
Specifically, the adaptive parameter $\gamma_k$ at time step $k$ is calculated as follows: 
\begin{align}
	\gamma_k &= \gamma_{\textrm{min}} + (\gamma_{\textrm{max}} - \gamma_{\textrm{min}}) r_k^2,
\end{align}
where the scalar $r_k = \frac{inlier}{total}$ is the ratio of inliers to the total number of radar points at time step $k$, and $\gamma_{\textrm{min}}$ and $\gamma_{\textrm{max}}$ are the pre-defined minimum and maximum values of $\gamma$, respectively.

Rather than directly tuning the adaptive parameter $\gamma$, we tune its minimum and maximum bounds.
This approach is more intuitive because there are relatively few logical ways to determine $\gamma_{\textrm{min}}$ and $\gamma_{\textrm{max}}$ without prior knowledge of the platform’s dynamic characteristics.
For instance, $\gamma_{\textrm{min}}$ could be chosen based on the radar’s velocity resolution, while $\gamma_{\textrm{max}}$ could be determined by the maximum expected velocity of the platform and its operating environment.
In indoor environments, where space is limited and speeds are generally lower, $\gamma_{\textrm{max}}$ may be tuned according to the platform’s maximum velocity.
Conversely, in outdoor settings, where platforms often operate at higher speeds, $\gamma_{\textrm{max}}$ can be set closer to the platform's true maximum velocity.

\begin{figure}[!t]
	\centering
	\includegraphics[scale=0.185]{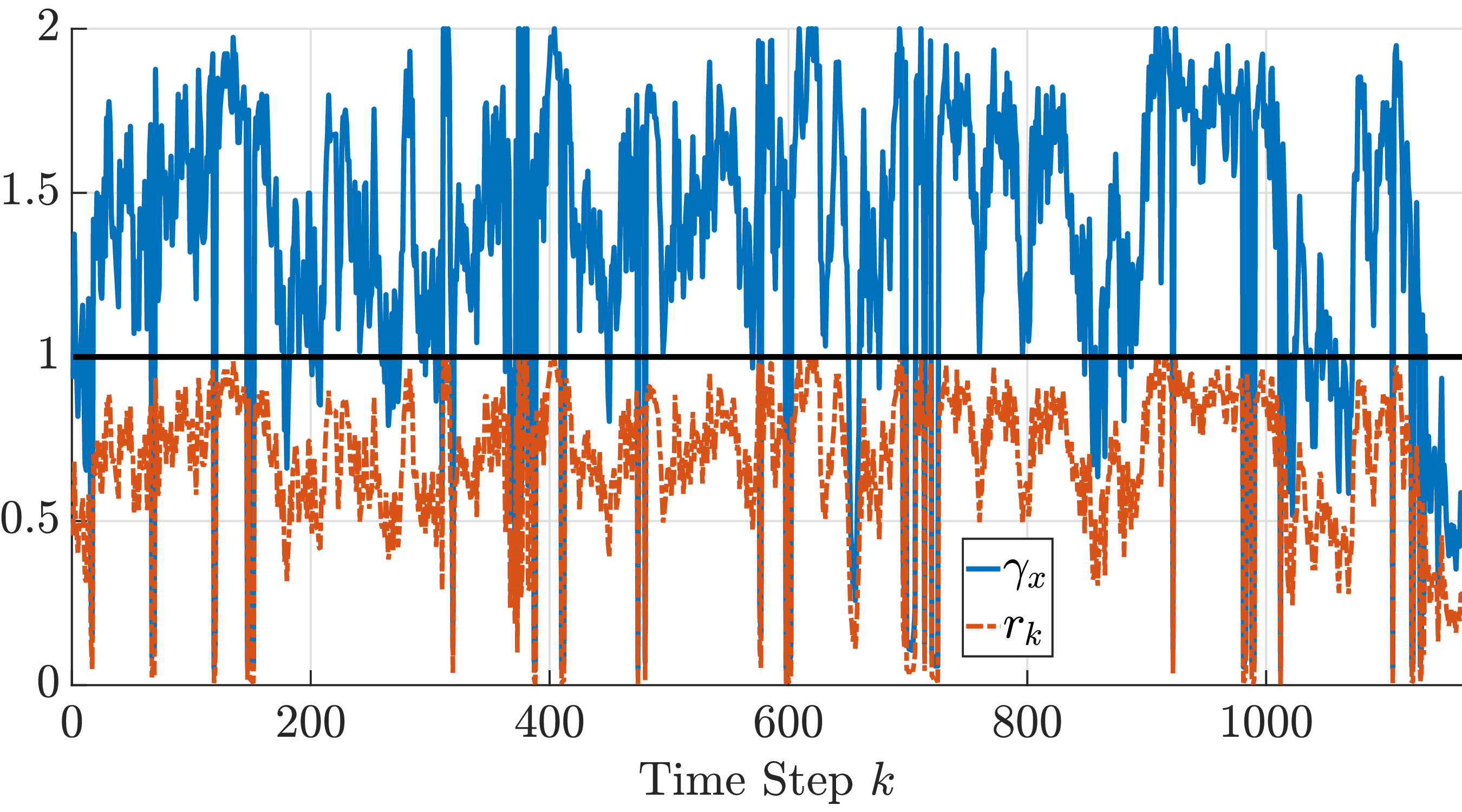}
	\caption{Relationship between the adaptive parameter $\gamma$ and the inlier ratio $r^2$. These results are drawn from the ColoRadar dataset (\texttt{arpg\_lab\_run1} trial), with $\gamma_{\textrm{min}} = \bbm 0.04 \!\!&\!\! 0.04 \!\!&\!\! 0.04 \ebm^\top$ m/s and $\gamma_{\textrm{max}} = \bbm 2 \!\!&\!\! 2 \!\!&\!\! 2 \ebm^\top$ m/s.}
	\label{gamma_ratio}
\end{figure}

Once the minimum and maximum bounds of $\gamma$ are established, its actual value for each radar scan is computed using the ratio $r$ with quadratic scaling.
The use of a quadratic function to map $\gamma$ to $r$ captures the nonlinear relationship between the confidence in the radar data and the required flexibility of the estimation. 
This quadratic mapping produces a sharper response when the ratio is close to one or zero, intentionally biasing $\gamma$ toward  $\gamma_{\textrm{min}}$ under conditions of low radar confidence and toward $\gamma_{\textrm{max}}$ when radar confidence is high.
As shown in Fig. \ref{gamma_ratio}, when the inlier ratio is low (indicating a high presence of outliers), $\gamma$ remains close to $\gamma_{\text{min}}$, enforcing stricter constraints on velocity estimation to limit the influence of erroneous radar data.
Conversely, with a high inlier ratio, $\gamma$ rises toward $\gamma_{\text{max}}$, permitting greater reliance on reliable radar measurements.
This adaptability is essential for maintaining the CREVE framework’s performance across diverse environmental conditions where radar data quality varies.
It is also worth noting that $\gamma$ can be interpreted as a weighting factor between the radar and accelerometer measurements, further underlining its importance in achieving an optimal fusion of sensor data.

\subsection{Accelerometer Bias Estimation}
In this subsection, we outline the process for estimating accelerometer bias using two consecutive radar ego-velocity measurements.
According to \cite{do2024dero}, the relationship between the estimated body velocity expressed in navigation frame $\hat{v}_b^n$ and the constrained ego-velocity $\tilde{v}^r$ can be established as
\begin{align} \label{radar_mechanization}
	\hat{v}_b^n &=  C_b^n C_r^b \tilde{v}^r - C_b^n \ssm{\bar{\omega}^b - \hat{b}_g} p_{r}^b,
\end{align}
where $\bar{\omega}^b$ is the raw gyroscope readings, $\hat{b}_g$ denotes gyroscope bias, and $p_r^b$ is the position of the radar relative to the $\{b\}$ frame.
In this study, we assume that $\hat{b}_g$, $p_r^b$, and $C_r^b$ are known.
These parameters can be obtained from a separated pre-processing step.
For example, a coarse alignment algorithm, involving approximately 10 seconds of stationary data, can be used to estimate $\hat{b}_g$, while $p_r^b$ and $C_r^b$ can be manually measured.

From this result and the previously described sensor model, one could yield the following equation
\begin{align} \label{ba_calc}
	\hat{b}_a = \bar{f}^b + C_n^b \Bigg( g^n - \dfrac{\hat{v}_{b, k}^n - \hat{v}_{b,k-1}^n}{\Delta t_r} \Bigg).
\end{align}
This calculation is practical and consistent with Eq. \eqref{a_cal} as described in Section \ref{subsection:lsq}.
The key difference is that, rather than compensating for bias and gravity to derive acceleration, \eqref{ba_calc} involves eliminating acceleration and gravity to directly estimate the bias.
Combining these two methods appears to create a coupling relationship.
However, we believe that by selecting $\gamma$ wisely, this coupling issue can be mitigated.
In other words, all aforementioned error sources can be captured by our proposed parameter $\gamma$, which make the system more robust against to these errors.
It is important to note that the bias estimation is influenced by the radar frequency, scaled by the factor $\Delta t_r$, which is typically around 10 Hz.
This frequency can lead to substantial bias fluctuations.
Since accelerometer bias varies slowly over time, we intentionally apply a low-pass filter to smooth the outcomes.
Fig. \ref{bias_estimation} illustrates the performance of the bias estimation process described in \eqref{ba_calc}.
The true accelerometer biases are estimated using an integrated system that fuses data from IMU and motion capture (Mocap) via an extended Kalman filter.
\subsection{Implementation}
\begin{figure}[!t]
	\centering
	\includegraphics[scale=0.185]{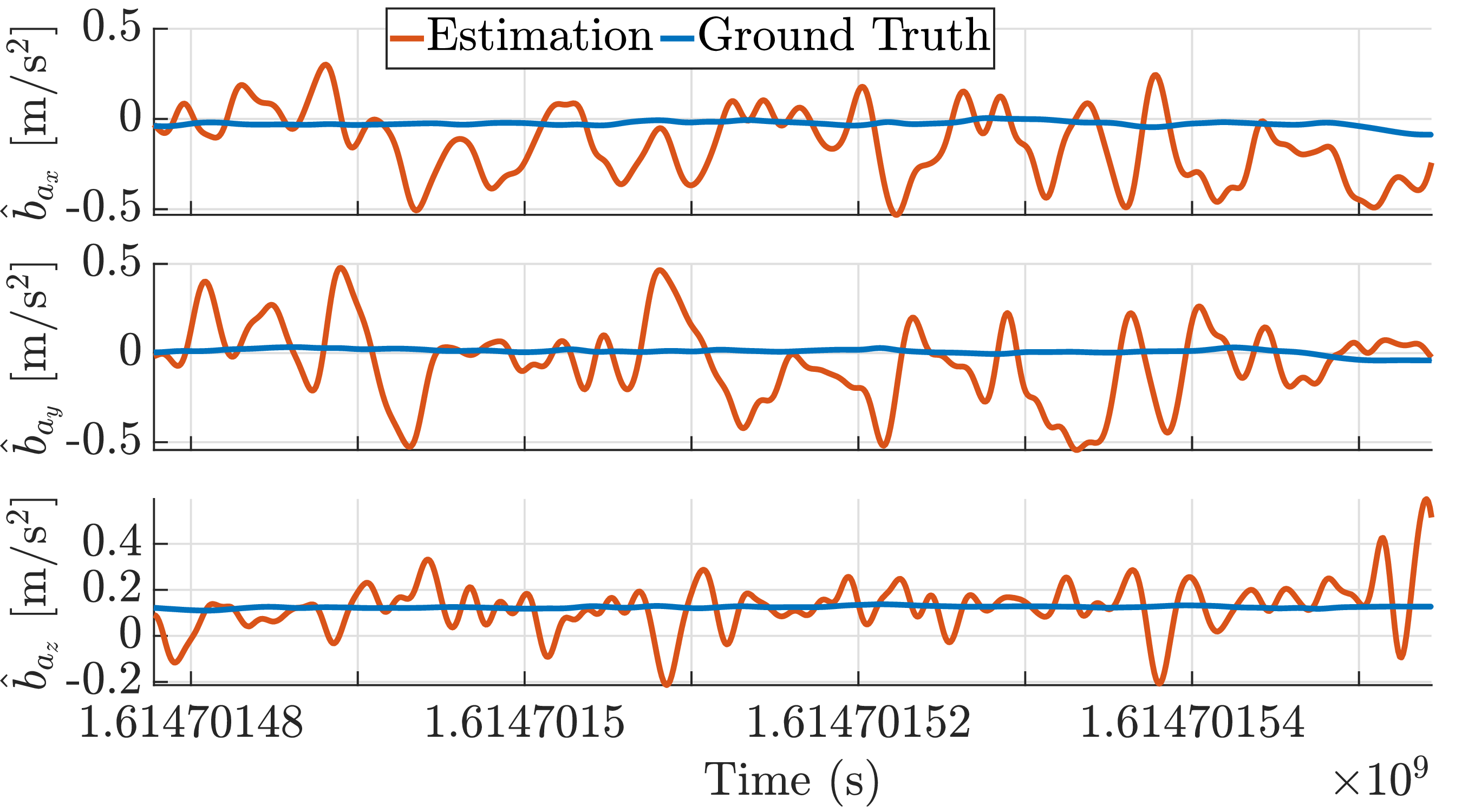}
	\caption{Accelerometer bias estimation of our CREVE, with a low-pass filter passband set to be 0.01 Hz.}
	\label{bias_estimation}
\end{figure}
In our implementation, we initiate the process with a brief stationary period (typically around 10 seconds) to perform a coarse alignment.
During this period, we compute an initial estimate of the gyroscope and accelerometer biases, which accelerates algorithm convergence and minimizes early-stage fluctuations.
We adopt zero-velocity detection as described in \cite{reve} by calculating the median of the Doppler velocity measurements from a set of radar point clouds.
If this median value falls below a predefined threshold $Z$, the prior ego-velocity estimate $\hat{v}_k^r$ is set to the zero vector.
Otherwise, the traditional RANSAC/LSQ is employed to estimate $\hat{v}_k^r$.
Subsequently, we estimate the sensor platform’s acceleration using Equation \eqref{ba_calc}, which is then used to establish the inequalities given in Equations \eqref{lp_1} and \eqref{lp_2}.
We check whether the prior estimate $\hat{v}_k^r$ satisfy these inequality.
satisfies these inequalities. If it does not, a constrained least squares approach is applied to yield the constrained radar ego-velocity estimation $\tilde{v}_k^r$.
Finally, this constrained estimation is used in the accelerometer bias calculation as outlined in Equation \eqref{bias_estimation}.
A detailed, step-by-step explanation of our CREVE is provided in Algorithm \ref{algo}.
\begin{algorithm}[!t]
	\caption{CREVE} \label{algo}
	\begin{algorithmic}[1]
		\State \textbf{Initialize:} $g^n$, $\gamma_{\textrm{max}}$, $\gamma_{\textrm{min}}$, zero-velocity threshold $Z$.
		\State Performing coarse alignment algorithm for stationary motion $\rightarrow \hat{b}_g, \hat{b}_a$.
		\For {$k = 1$ to $K$}
			\State Computing the median $n$ of $v_{D,i}$ in $\mathcal{P}_k$.
		\If {$n < Z$}
			\State Assigning $\hat{v}_k^r = 0$.
		\Else
			\State Performing RANSAC/LSQ $\rightarrow \hat{v}_k^r$.
		\EndIf
		\State Computing acceleration $\leftarrow$ \eqref{a_cal}.
		\If {$\hat{v}_k^r$ does not satisfy \eqref{lp_1} and \eqref{lp_2}}
			\State Solving the LP \eqref{lp} $\rightarrow \tilde{v}_k^r$.
			\State Calculating accelerometer bias $\leftarrow$ \eqref{ba_calc}.
		\Else
			\State $\tilde{v}_k^r = \hat{v}_k^r$.
		\EndIf
		\EndFor
	\end{algorithmic}
\end{algorithm}
\section{Evaluation} \label{sec:results}
\subsection{Open-source datasets}
To validate the proposed method (\texttt{Ours}), we conducted evaluations using two well-established open-source datasets: the IRS dataset \cite{DoerIros2021} and the ColoRadar dataset \cite{coloRadar}.
These datasets were intentionally selected due to their diversity and comprehensive coverage of real-world indoor scenarios, which enable a rigorous assessment of the method’s performance under a wide range of operational conditions.
The IRS dataset comprises five distinct trials: \texttt{easy}, \texttt{medium}, \texttt{difficult}, \texttt{dark}, and \texttt{dark fast}.
Each trials was conducted using a MoCap system to generate high-precision ground truth data.
The scenarios cover a wide range of motion dynamics, from slow to rapid movement, thereby providing a rigorous test of the robustness and adaptability of the evaluated methods.
On the other hand, we selected four experiments from the ColoRadar dataset: \texttt{arpg\_lab\_run1}, \texttt{ec\_hallways\_run1}, \texttt{longboard\_run1}, and \texttt{edgar\_army\_run5}.
These experiments cover a diverse set of indoor environments, including laboratory spaces, hallways and tunnels.
Ground truth for this dataset was obtained through a globally optimized pose graph that integrates information from an inertial measurement unit (IMU), LiDAR, and loop closure constraints.

A concise overview of the data collection setups is provided here, while further details can be found in the respective publications.
The IRS dataset was recorded in a small indoor laboratory using a drone-mounted sensor platform equipped with a 4D FMCW TI IWR6843AOP radar and an Analog Devices ADIS16448 IMU.
The IMU operated at approximately 400 Hz, while the radar produced 4D point clouds at a frequency of 10 Hz, offering a 120-degree field of view in both azimuth and elevation.
Meanwhile, the ColoRadar dataset employed a ground-based sensor platform equipped with a Lord Microstrain 3DM-GX5-15 IMU, operating at 300 Hz, and a 77 GHz single-chip TI AWR1843BOOST-EVM radar module paired with a DCA1000-EVM, capturing radar data at 10 Hz.

\subsection{Comparative Method}
We implemented our CREVE framework using MATLAB R2022b, running on a system equipped with an Intel i9-12900K CPU operating at 3.20 GHz.
The proposed method (\texttt{Ours}) is compared with the following three state-of-the-art algorithms:
\begin{itemize}
	\item \textbf{REVE} \cite{reve}: A standard RANSAC/LSQ-based framework for ego-velocity estimation.
	\item \textbf{DeREVE} \cite{dreve}: A RANSAC/LSQ-based decoupling method designed to enhance the accuracy of velocity estimation along the $z$-axis.
	\item \textbf{RAVE} \cite{rave}: An extension of the REVE framework that incorporates mechanisms for detecting and rejecting anomalies in velocity estimation.
\end{itemize}

The parameters used in our experiments are as follows: the gravitational constant is set to $g = 9.81$ m/s$^2$, and the zero-velocity threshold to $Z = 0.05$.
For the RANSAC algorithm, the success probability is set to 0.99, and the outlier probability to 0.4.
Regarding the adaptive parameter settings, for the IRS dataset, we configure $\gamma_{\textrm{min}} = \bbm 0.01 \!\!&\!\! 0.01 \!\!&\!\! 0.01 \ebm^\top$ for all experiments, based on the radar's velocity resolution.
For $\gamma_{\textrm{max}}$, we use $\bbm 1.25 \!\!&\!\! 1.25 \!\!&\!\! 1.25 \ebm^\top$ m/s in the Mocap \texttt{Difficult} and \texttt{Dark Fast} trials, and $\bbm 0.75 \!\!&\!\! 0.75 \!\!&\!\! 0.75 \ebm^\top$ m/s for the remaining trials.
These values are selected based on the radar configuration.
Specifically, $\gamma_{\textrm{min}}$ is determined with respect to the Doppler velocity resolution, while $\gamma_{\textrm{max}}$ is set according to the operational maximum velocity, taking into account that the drone operates within a confined indoor environment.
For the \texttt{Difficult} and \texttt{Dark Fast} trials, we increase $\gamma_{\textrm{max}}$ to account for the drone's higher motion speed.
Following the same rationale, for all trials in the ColoRadar dataset, we set $\gamma_{\textrm{min}} = \bbm 0.04 \!\!&\!\! 0.04 \!\!&\!\! 0.04 \ebm^\top$ m/s and $\gamma_{\textrm{max}} = \bbm 2 \!\!&\!\! 2 \!\!&\!\! 2 \ebm^\top$ m/s, based on the reported velocity resolution and maximum limits in \cite{coloRadar}.

For the evaluation benchmark, we employ two complementary error metrics.
First, we quantify the ego-velocity estimation accuracy using the time‑averaged root‑mean‑square error (RMSE) for each axis.
For $i = \{1,2,3\}$, corresponding to the $x,y$ and $z$ directions, the RMSE is defined as $\textrm{RMSE}(v^r_{i}) = \sqrt{\frac{1}{K} \sum_{k=1}^K ( \hat{v}^{r}_{i,k} - v^{r,gt}_{i,k} )^2}$, where $K$ is the total number of time steps, $\hat{v}^{r}_{i,k}$ is the estimated ego-velocity along axis $i$ at time step $k$, and $v^{rg,t}_{i,k}$ denotes the corresponding ground‑truth ego‑velocity.
Second, we evaluate long‑term consistency using the absolute trajectory error (ATE), which quantifies the cumulative drift of the odometry estimate relative to the ground‑truth path.
The ATE has become the de facto standard in the robotics and SLAM communities for benchmarking odometry and mapping systems.
This makes it straightforward to compare performance across different methods and datasets.
To generate the odometry trajectories used in the ATE computation, we numerically integrate $\eqref{radar_mechanization}$.
Denoting by $p_{b,0}^n$ the known initial position at time step $k=0$, we propagate the position as
\begin{align}
	p_{b,k}^n = p_{b,k-1}^n + \hat{v}_b^n \Delta t_r,
\end{align}
where the rotation matrix $C_{b,k}^n$ in $\eqref{radar_mechanization}$ is computed directly from the dataset's ground-truth orientation (e.g., provided as quaternions).

\subsection{Analysis}
\begin{figure}[!t]
	\centering
	\includegraphics[scale=0.185]{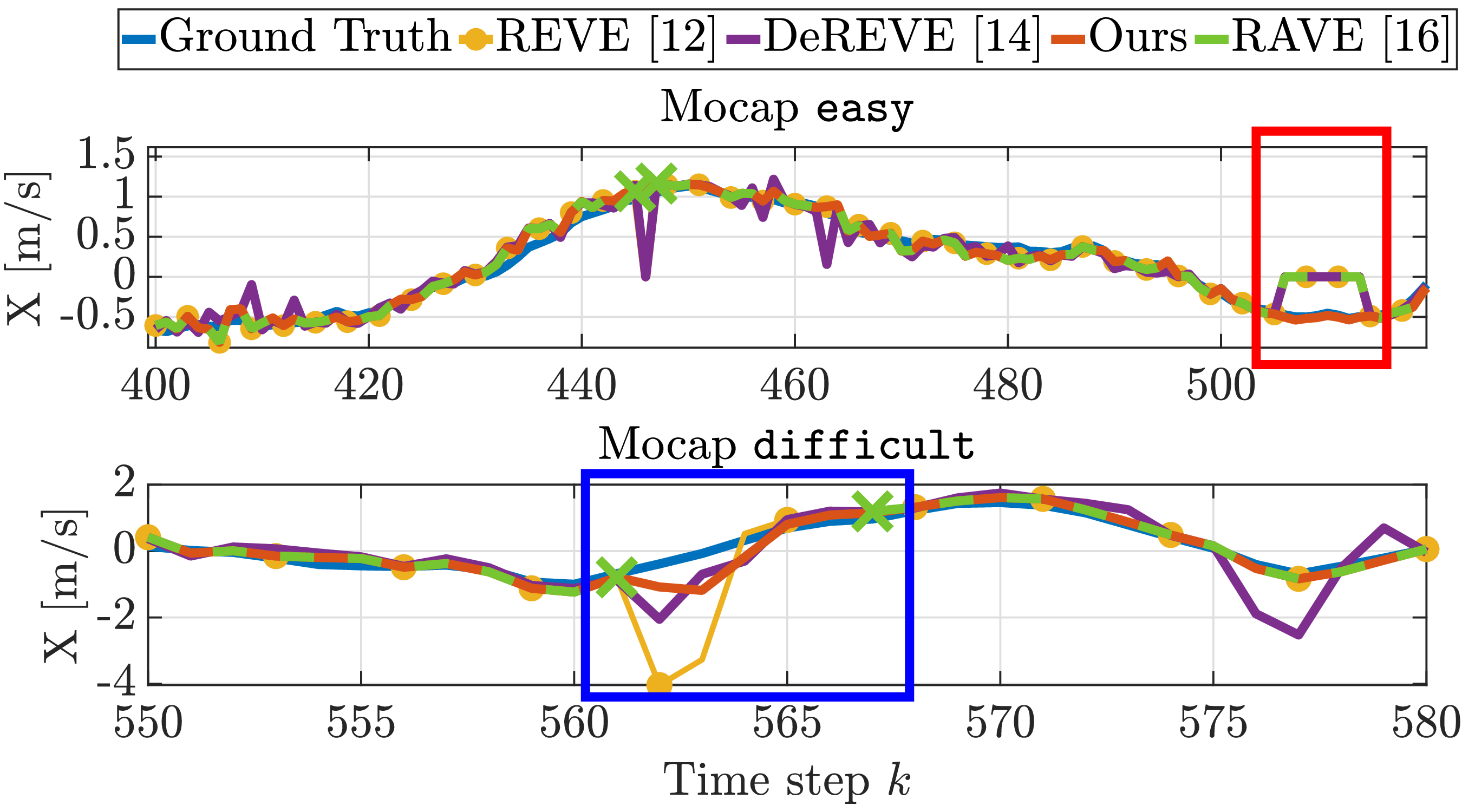}
	\caption{Comparison of ego-velocity estimation results across four investigated approaches using the MoCap \texttt{easy} and \texttt{difficult} datasets (best viewed in color).}
	\label{velocity_estimation}
\end{figure}
\begin{figure*}[!t]
	\subfloat[Bar chart of ATE statistical across methods.]{\label{Fig6:a} \includegraphics[width=0.30\linewidth, keepaspectratio]{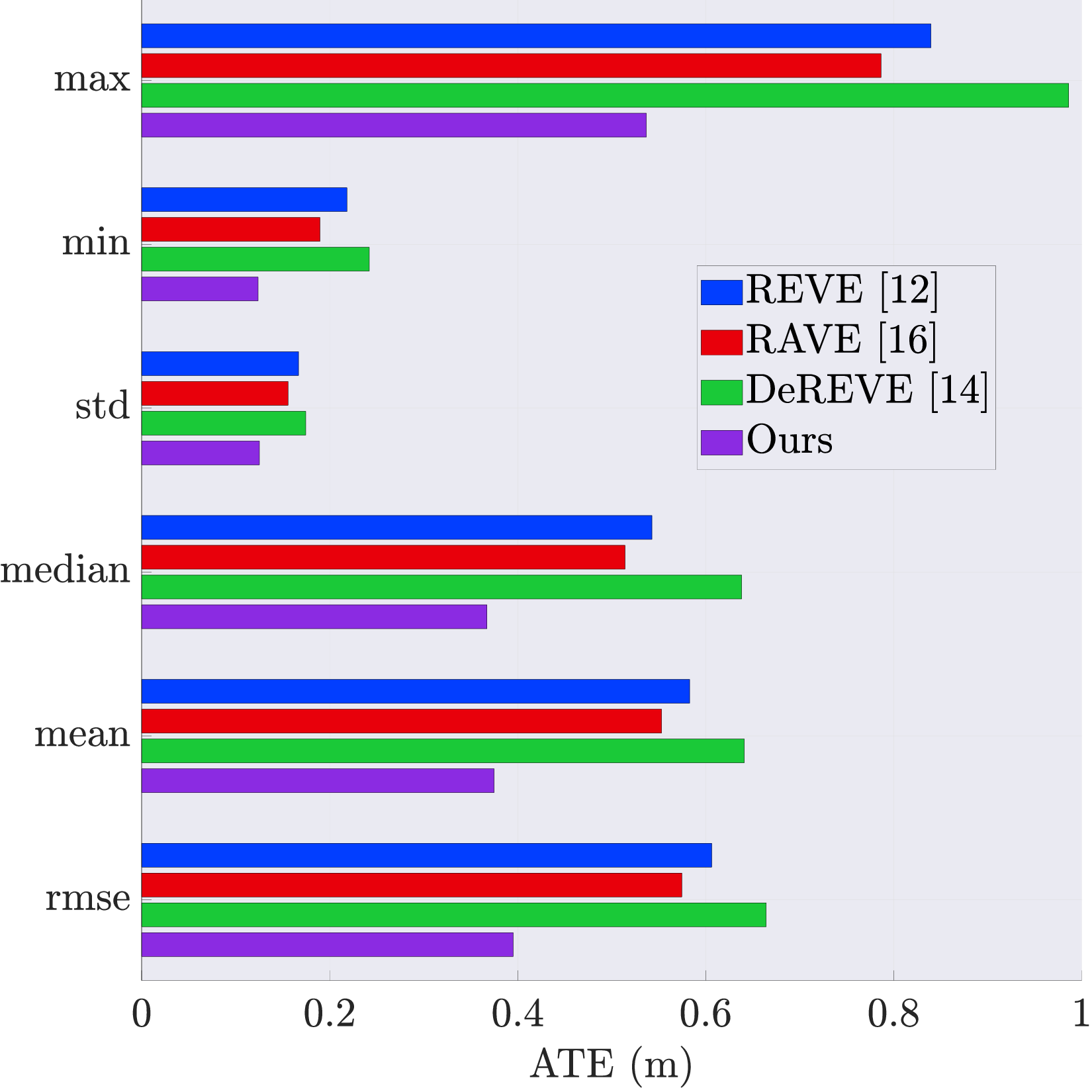}}%
	\hfill
	\subfloat[ATE comparison over time.]{\label{Fig6:b} \includegraphics[width=0.32\linewidth, keepaspectratio]{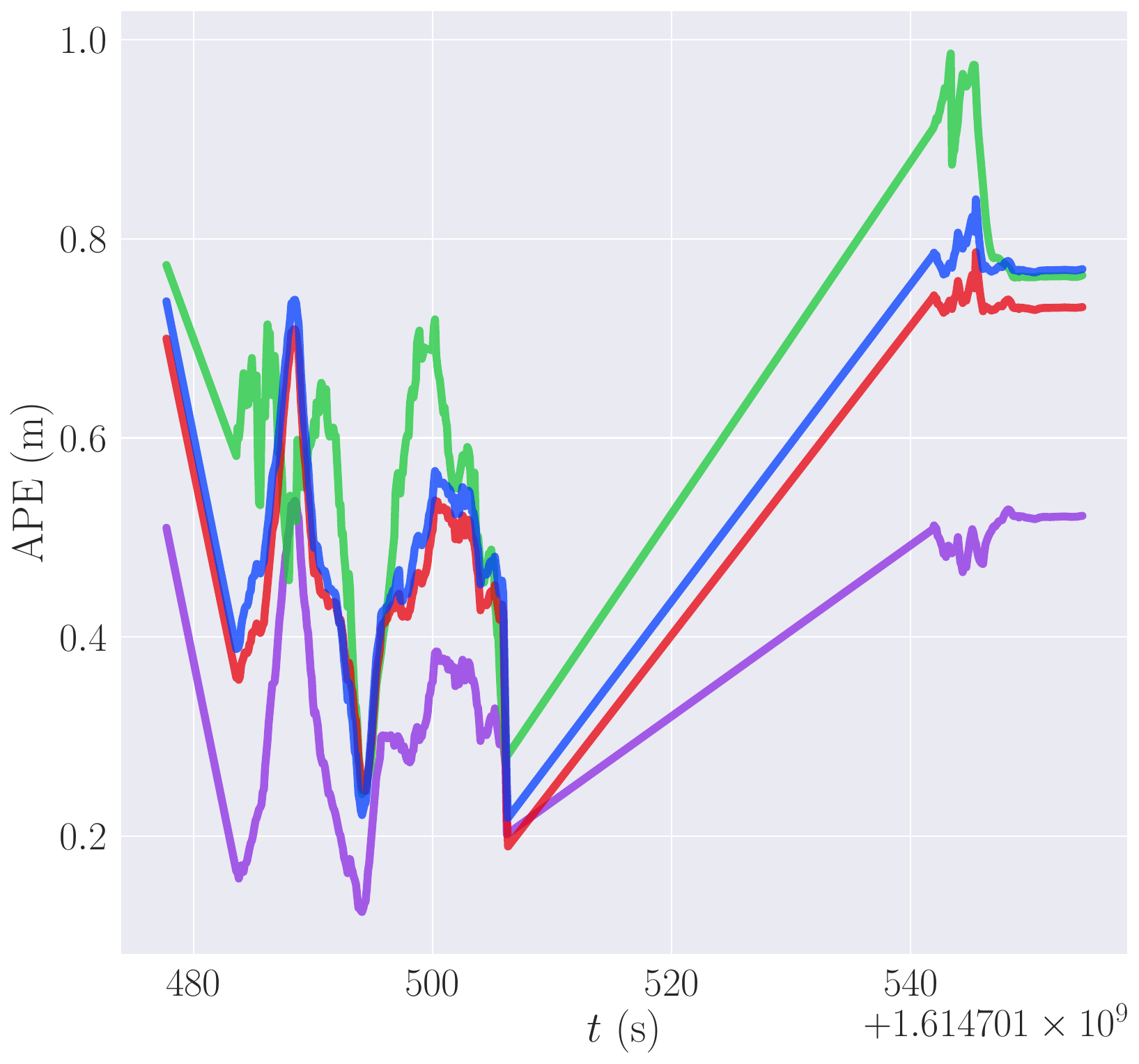}}%
	\hfill
	\subfloat[2D trajectory estimation in the $x$-$y$ plane for the MoCap \texttt{easy} trial, with ``$\times$'' marking the end of the trajectory.]{\label{Fig6:c} \includegraphics[width=0.258\linewidth, keepaspectratio]{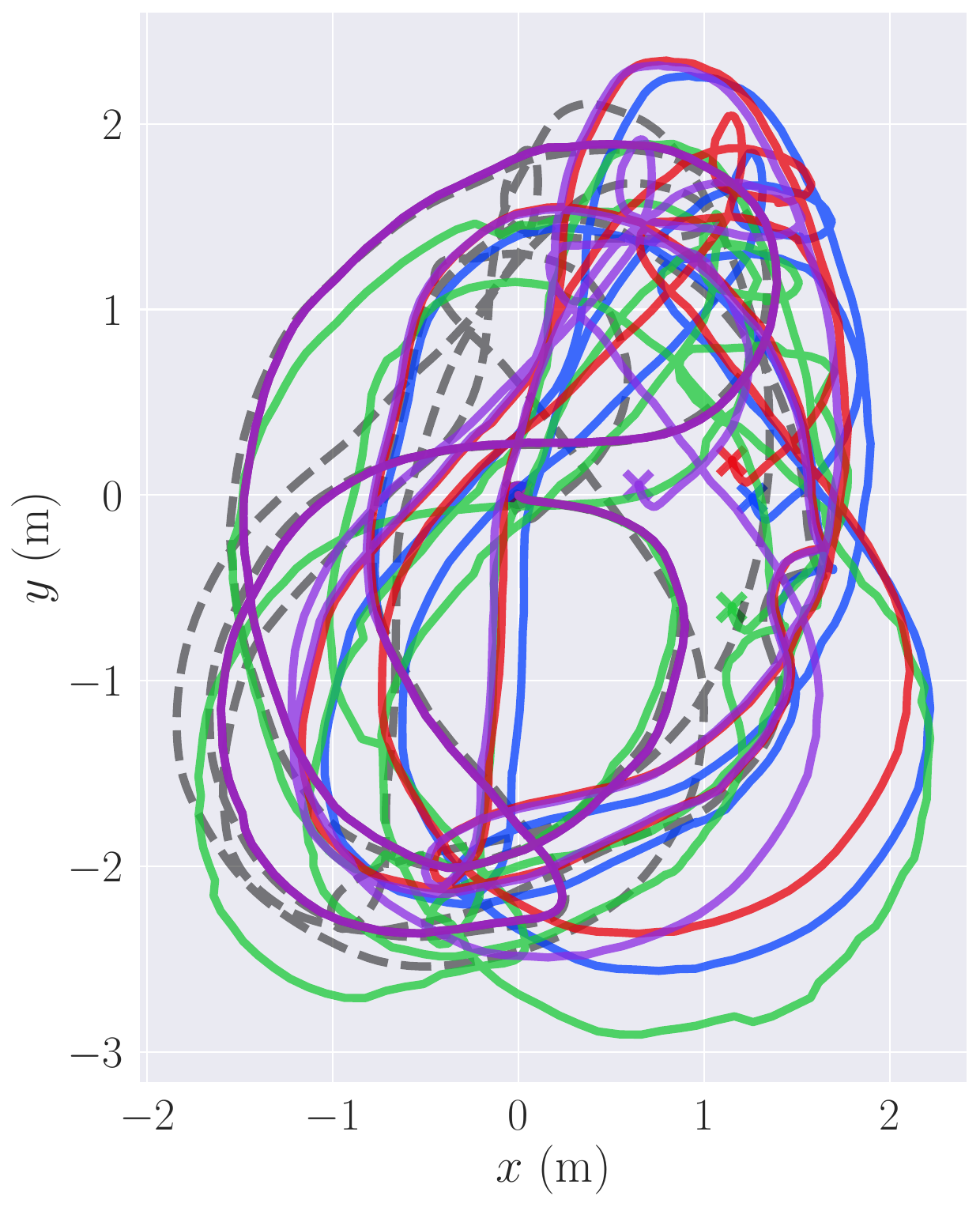}}
	\hfill
	\subfloat[Position estimation along the $x$, $y$, and $z$ axes for the MoCap \texttt{easy} trial.]{\label{Fig6:d} \includegraphics[width=0.315\linewidth, keepaspectratio]{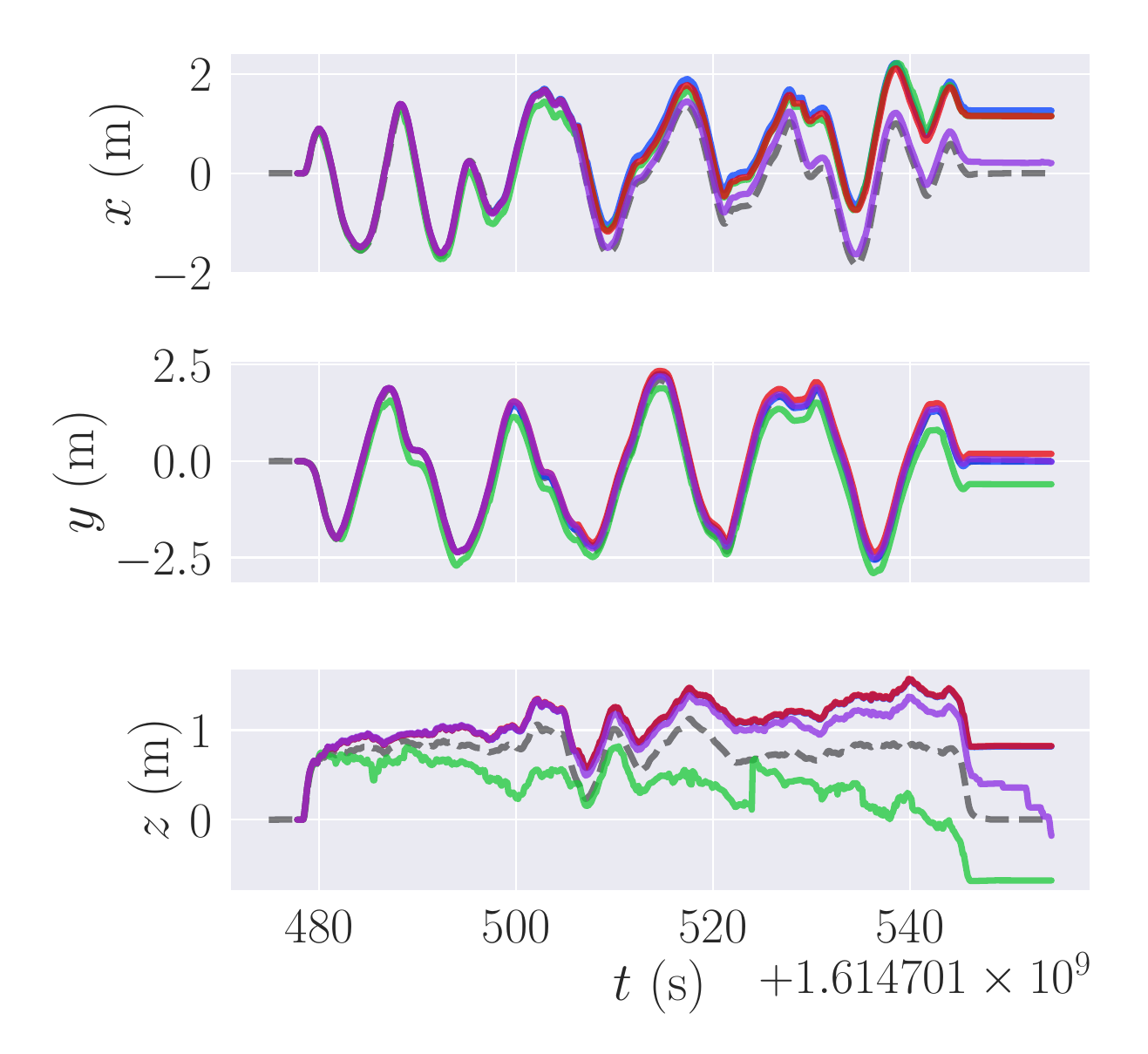}}%
	\hfill
	\subfloat[Position estimation along the $x$, $y$, and $z$ axes for the MoCap \texttt{medium} trial.]{\label{Fig6:e} \includegraphics[width=0.32\linewidth, keepaspectratio]{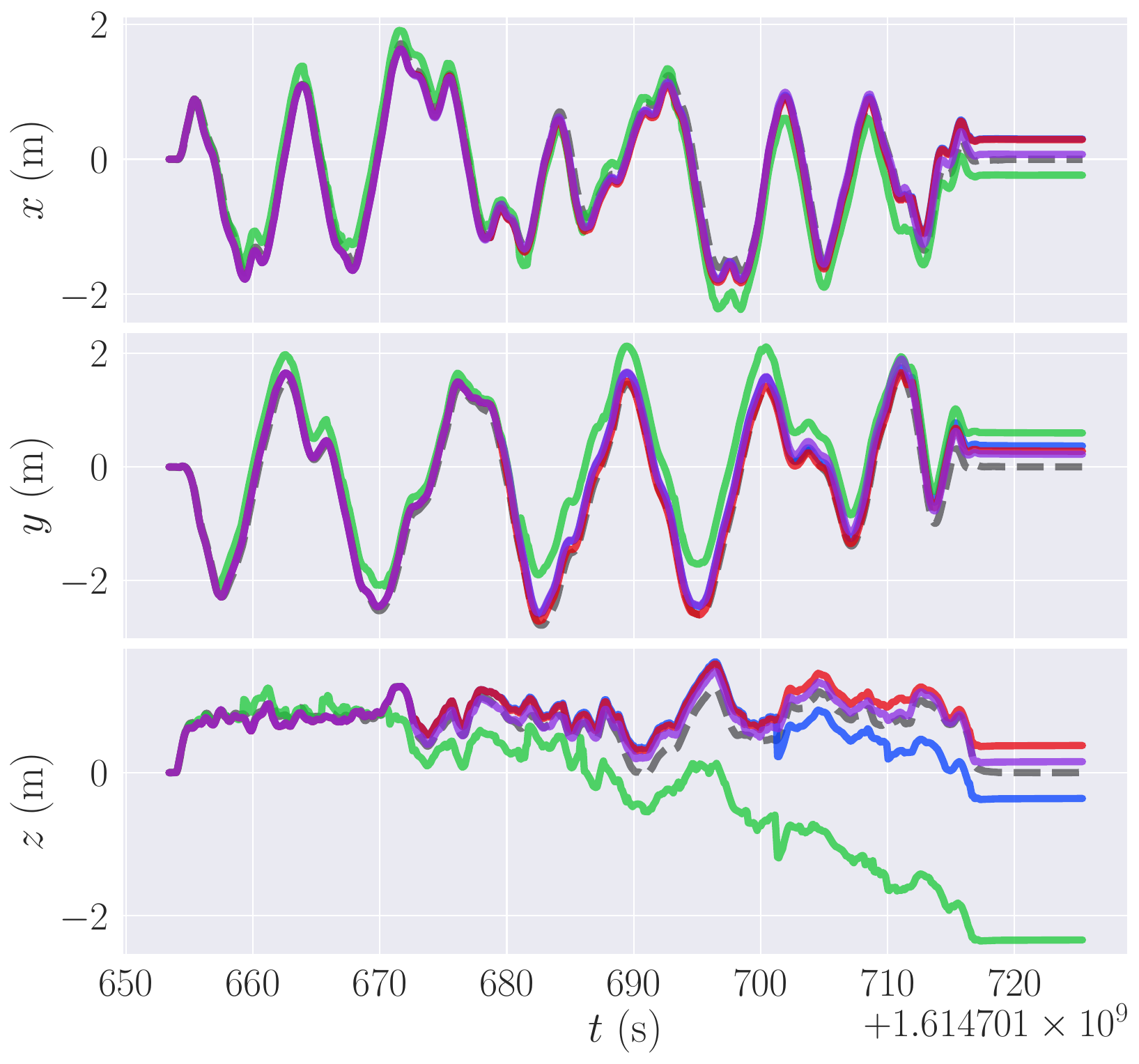}}%
	\hfill
	\subfloat[Position estimation along the $x$, $y$, and $z$ axes for the MoCap \texttt{difficult} trial.]{\label{Fig6:f} \includegraphics[width=0.32\linewidth, keepaspectratio]{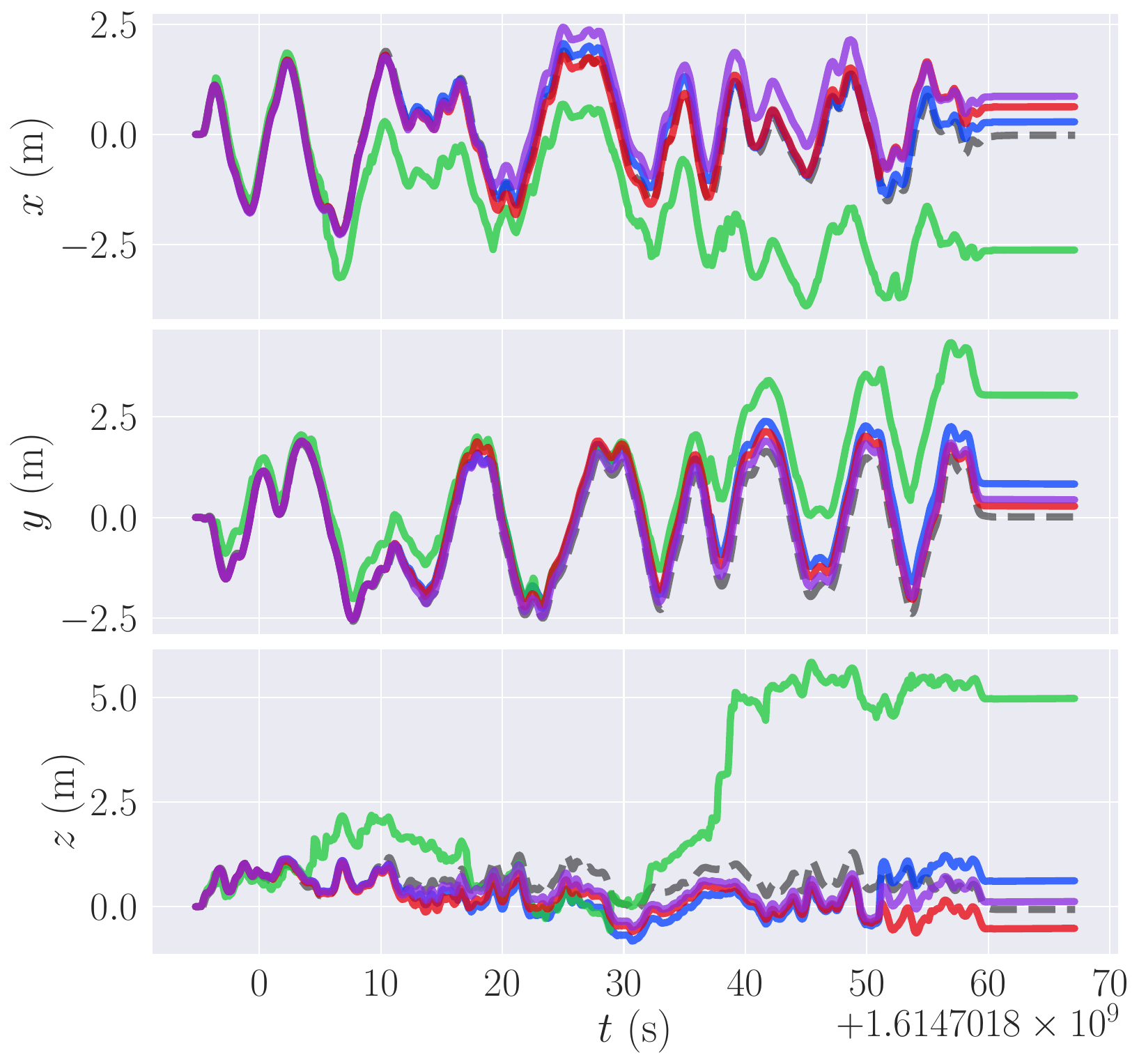}}
	\caption{
		Odometry estimation of the four methods compared to the MoCap ground truth (dash line) using the toolbox \cite{grupp2017evo} using Umeyama alignment method (best viewed in color).
	}
	\label{Fig6}
\end{figure*} 
\renewcommand{\arraystretch}{1.1}
\begin{table*}[!t]
	\caption{Comparative evaluation of four 3D ego‑velocity estimation methods on the IRS \cite{DoerIros2021} and ColoRadar \cite{coloRadar} datasets. Reported metrics include time‑averaged RMSE along the $x, y$, and $z$ axes, mean computation time per dataset, and the RAVE method’s radar‑scan rejection rate.}
	\label{table1}
	\begin{center}
		\begin{tabular}{>{\centering\arraybackslash}p{2.5cm} >{\centering\arraybackslash}p{0.7cm} |>{\centering\arraybackslash}p{0.7cm} | >{\centering\arraybackslash}p{0.7cm} >{\centering\arraybackslash}p{0.8cm} |>{\centering\arraybackslash}p{0.8cm} | >{\centering\arraybackslash}p{0.8cm} >{\centering\arraybackslash}p{0.7cm} |>{\centering\arraybackslash}p{0.7cm} | >{\centering\arraybackslash}p{0.7cm} | >{\centering\arraybackslash}p{0.85cm} >{\centering\arraybackslash}p{0.7cm} |>{\centering\arraybackslash}p{0.7cm} | >{\centering\arraybackslash}p{0.7cm}}
			\toprule[1pt]
			
			\multirow{2}{*}{Datasets} & \multicolumn{3}{c}{REVE \cite{reve} (\textbf{0.007} s/loop)} & \multicolumn{3}{c}{DeREVE \cite{dreve} (0.013 s/loop)}  & \multicolumn{4}{c}{RAVE \cite{rave} (\textbf{0.007} s/loop)} & \multicolumn{3}{c}{\texttt{Ours} ({0.008} s/loop)} \\
			
			\cmidrule(lr){2-4} \cmidrule(lr){5-7} \cmidrule(lr){8-11} \cmidrule(lr){12-14}
			
			&  $v_x$  &  $v_y$  &  $v_z$  &  $v_x$  &  $v_y$  &  $v_z$  &  $v_x$  &  $v_y$  &  $v_z$  & Reject&  $v_x$  &  $v_y$  &  $v_z$  \\
			
			\cmidrule(lr){1-1} \cmidrule(lr){2-4} \cmidrule(lr){5-7} \cmidrule(lr){8-11} \cmidrule(lr){12-14}
			
			\texttt{Easy}        &  0.120  &  0.069  & \textbf {0.086}  &  0.178  &  0.111  &  0.292  &  {0.103}  &  {0.067}  &  \textbf{0.086}  & 0.52\%  &  \textbf{0.081}  & \textbf{0.040}  &  {0.088}  \\
			\texttt{Medium}  &  0.158  &  0.085  &  0.262  &  0.310  &  0.178  &  0.350  &  {0.155}  &  {0.078}  &  {0.128} & 0.97\%  &  \textbf{0.131}  & \textbf {0.072}  &  \textbf{0.109}  \\
			\texttt{Difficult}   &  0.307  &  0.269  &  0.319 &  0.652  &  0.490  &  0.840  &  {0.151}  &  {0.101}  &  {0.178}  & 4.16\%  &  \textbf{0.150}  &  \textbf{0.097}  &  \textbf{0.177}  \\
			\texttt{Dark}        &  0.153  &  0.080  &  0.132  &  0.389  &  0.177  &  0.449   &  {0.138}  &  {0.079}  &  {0.122}  & 0.63\%  &  \textbf{0.135}  &  \textbf{0.078}  & \textbf{0.121}  \\
			\texttt{Dark fast}  &  0.197  &  0.248  &  0.196  &  0.517  &  0.478  &  0.832  &  {0.117}  &  \textbf{0.076}  &  {0.129}  & 1.24\%  &  \textbf{0.108}  & {0.117}  & \textbf{0.125}  \\
			\cmidrule(lr){1-1} \cmidrule(lr){2-4} \cmidrule(lr){5-7} \cmidrule(lr){8-11} \cmidrule(lr){12-14}
			\texttt{arpg\_lab\_run1}        &  0.180  &  0.176  &  0.182  &  0.268  &  0.268  &  0.330  & 0.167  &  0.170  &  0.177 & 0.86\%  & \textbf{0.140}  & \textbf{0.161}  &  \textbf{0.173}  \\
			\texttt{ec\_hallways\_run1}   &  0.230  &  0.223  &  0.268  &  0.376  &  0.373  &  0.547  &  0.208  & \textbf{0.219}  & \textbf{0.267} & 0.50\%  &  \textbf{0.202}  & \textbf{0.219}  &  \textbf{0.267}  \\
			\texttt{longboard\_run1}     &  1.279  &  4.418  &  0.633 &  0.997  &  4.848  &  0.461  &  1.191  &  \textbf{4.174}  &  0.545 & 24.54\%  &  \textbf{1.110}  &  4.197  & \textbf{0.493}  \\
			\texttt{edgar\_army\_run5}   &  0.241  &  0.171  &  0.282  &  0.381  &  0.297  &  0.419   &  \textbf{0.228}  &  0.160  &  0.275 &  0.91\% & 0.229  &  \textbf{0.155}  & \textbf{0.274}  \\
			
			\cmidrule(lr){1-1} \cmidrule(lr){2-4} \cmidrule(lr){5-7} \cmidrule(lr){8-11} \cmidrule(lr){12-14}
			
			Mean   &  0.318  &  0.638 &  0.262  &  0.452  &  0.801  &  0.502  &  {0.273}  &  {0.568} &  {0.212}  & 3.81\% &  \textbf{0.253}  &  \textbf{0.461}  &  \textbf{0.202}  \\
			
			\bottomrule[1pt]
			\multicolumn{14}{l}{\footnotesize Bold numbers indicate the best results (smallest values) and all values are rounded to three decimals.}
		\end{tabular}
	\end{center}
\end{table*}
In this subsection, we provide an in-depth analysis of the performance differences observed among the four compared methods across diverse scenarios and benchmarks.
\subsubsection{Velocity Estimation}
The $v^r$ estimation performance of all techniques on the Mocap \texttt{easy} and \texttt{difficult} datasets is highlighted in Fig. \ref{velocity_estimation}.
In typical scenarios, all of them exhibit comparable performance, with the exception of DeREVE, which displays some random peaks in its estimates.
In extreme case (the blue box area) where REVE reveals significant estimation errors, CREVE maintains robust performance.
In contrast, RAVE consistently rejects anomalous data, but at the cost of producing discontinuous estimates.
Notably, $\texttt{Ours}$ also shows robustness against incorrect zero-velocity detection, as evidenced by the red box in the MoCap \texttt{easy} scenario.
This robustness arises from the CREVE's implementation, as described in Algorithm \ref{algo}, where the output of the zero-velocity detection module is further validated through the proposed inequality constraint.
If this constraint is not satisfied, indicating a conflict between radar-based estimation and IMU measurements, CREVE autonomously corrects the detection failure.
\begin{table}[!t]
	\caption{Absolute trajectory error calculated with \textit{pos-yaw} alignment \cite{eval}, presented in terms of translation errors across five trials of the IRS \cite{DoerIros2021} dataset for all considered approaches.}
	\label{table2}
	\begin{center}
		\begin{tabular}{c >{\centering\arraybackslash}p{1.15cm}|>{\centering\arraybackslash}p{1.15cm}|>{\centering\arraybackslash}p{1.15cm}|>{\centering\arraybackslash}p{1.15cm}}
			\toprule[1pt]
			Datasets & REVE \cite{reve} & DeREVE \cite{dreve} & RAVE \cite{rave} & \texttt{Ours} \\
			
			\cmidrule(lr){1-1} \cmidrule(lr){2-5}
			\texttt{Easy}       & 0.597 & 0.563 & {0.563} & \textbf{0.378} \\
			\texttt{Medium} & 0.356 & 1.045 & {0.260} & \textbf{0.162} \\
			\texttt{Difficult} & 0.649 & 2.738 & {0.483} & \textbf{0.478} \\
			\texttt{Dark}      & 0.342 & 0.968 & {0.292} & \textbf{0.271} \\
			\texttt{Dark Fast} & 0.529 & 2.003 & \textbf{0.206} & {0.292} \\
			\cmidrule(lr){1-1} \cmidrule(lr){2-5}
			Mean ATE (m) & 0.495 & 1.463 & {0.361} & \textbf{0.316} \\
			\bottomrule[1pt]
			\multicolumn{5}{l}{\footnotesize The bold numbers represent the best results (smallest numbers).} \\
			\multicolumn{5}{l}{\footnotesize All values are rounded to three decimal digits.}
		\end{tabular}
	\end{center}
\end{table}
\subsubsection{Velocity Estimation Root Mean Square Error}
Table \ref{table1} presents the ego-velocity RMSE results for both the IRS and ColoRadar datasets.
Overall, the proposed CREVE method consistently outperforms all other approaches across the evaluated scenarios. In comparison to the conventional REVE framework, CREVE achieves an average RMSE reduction of approximately 20\% along the $x$-axis, 28\% along the $y$-axis, and 23\% along the $z$-axis.
Conversely, DeREVE exhibits the lowest estimation accuracy, while RAVE performs comparably to CREVE in terms of RMSE.
However, RAVE discards a substantial portion of radar point cloud data, particularly in the \texttt{longboard\_run1} trial, where 408 out of 1662 radar frames were rejected, corresponding to a rejection rate of 24.54\%.
Although RAVE demonstrates improved accuracy relative to REVE, its high rejection rate presents a significant limitation.
In contrast, CREVE not only achieves superior estimation accuracy but also addresses this drawback by effectively utilizing more radar data without compromising robustness.
\subsubsection{Absolute Trajectory Error}
The ATE results, computed using established toolboxes \cite{grupp2017evo, eval} and presented in Table II and Fig. 7, further validate CREVE’s superior performance.
On average, CREVE achieves an ATE of 0.316 m, compared to 0.495 m for REVE, 1.463 m for DeREVE, and 0.361 m for RAVE, representing reductions of approximately 36\%, 78\%, and 12\%, respectively.
Figure \ref{Fig6:c} depicts the odometry trajectories for the \texttt{easy} dataset, where CREVE’s path closely aligns with the ground truth, while REVE and DeREVE exhibit noticeable drift.
Notably, all trajectories are initiated and terminated at the coordinate origin; as shown in the plot, CREVE’s endpoint (denoted by an ``$\times$” marker) lies closer to the origin than the endpoints produced by any of the other methods.
The results in Fig. \ref{Fig6:b} and Fig. \ref{Fig6:a} reveals that CREVE achieves the lowest median, minimum, and maximum ATE, along with the smallest variance, demonstrating superior accuracy and consistency.
This performance is attributable to CREVE’s constrained velocity estimates, which minimize cumulative errors over time, making it particularly suitable for long-term navigation tasks.

\subsubsection{Computation Time}
The average computation time per radar point cloud set, reported in Table \ref{table2}, indicates that CREVE’s processing time (0.008 s/loop) is comparable to REVE (0.007 s/loop) and RAVE (0.007 s/loop), while DeREVE requires significantly longer (0.013 s/loop) due to its double RANSAC/LSQ structure.
CREVE’s efficiency, despite incorporating additional constraints and bias estimation, ensures its feasibility for real-time applications.
This balance between computational cost and performance enhancement positions CREVE as a practical solution for radar-inertial navigation systems requiring both accuracy and responsiveness.

\section{Conclusion} \label{sec:conclusion}
In this study, we have proposed CREVE, an innovative acceleration-based constraint approach for robust radar ego-velocity estimation, designed to enhance RIO systems.
By integrating IMU-derived acceleration measurements as inequality constraints, CREVE bounds radar-based velocity estimates within physically feasible limits, effectively mitigating the impact of outliers prevalent in sparse point clouds from mmWave FMCW radars.
The proposed adaptation rule for the constraint parameter $\gamma$, driven by the inlier ratio of the radar point cloud, dynamically adjusts the balance between radar and inertial data, ensuring adaptability across diverse operational conditions.
Additionally, the incorporation of accelerometer bias estimation, utilizing consecutive constrained velocity estimates, maintains the accuracy of acceleration data over extended periods, further bolstering system reliability.
Quantitative evaluations on the IRS and ColoRadar datasets demonstrate CREVE’s superior performance over state-of-the-art methods, including REVE, DeREVE, and RAVE.
The proposed method achieves significant reductions in RMSE for ego-velocity estimation, while maintaining computational efficiency comparable to baseline approaches.
These results highlight CREVE’s robustness in challenging scenarios, such as dynamic indoor environments, and its ability to provide continuous, accurate velocity estimates without discarding measurements.

CREVE represents a substantial advancement in radar-inertial navigation, offering a reliable submodule for RIO systems where traditional radar-only methods falter.
The strong synergy between radar and IMU, explicitly demonstrated through this framework, paves the way for improved odometry in robotics applications.
Future work will investigate the integration of CREVE as a submodule within our recent RIO framework, DeRO \cite{do2024dero}, to further enhance system performance and robustness.

%\section*{Acknowledgments}
%This should be a simple paragraph before the References to thank those individuals and institutions who have supported your work on this article.
\bibliographystyle{IEEEtran}
\bibliography{IEEEabrv,ieee_tim_viet}
%


%\bf{If you include a photo:}\vspace{-33pt}


%\vfill

\end{document}